%% file: 0_main.tex
\documentclass[letterpaper, 10 pt, conference]{ieeeconf} 

\IEEEoverridecommandlockouts                              

\usepackage{graphics} 
\usepackage{epsfig} 
\usepackage{mathptmx} 
\usepackage{times} 
\usepackage{amsmath} 
\usepackage{amssymb}  
\usepackage{xcolor}
\usepackage{float}

\usepackage{graphicx} 
\usepackage{mathptmx} 
\usepackage{amsmath} 
\usepackage{amssymb}  
\usepackage{amsfonts}
\usepackage{listings}
\usepackage{verbatim}
\usepackage{color}
\usepackage{soul}
\usepackage{wrapfig}
\usepackage{cite}
\usepackage[labelformat=simple]{subcaption}
\usepackage{balance}
\usepackage[bordercolor=white,backgroundcolor=gray!30,linecolor=black,colorinlistoftodos]{todonotes}

\author{Aaron Weber$^{1}$, Daksh Dhingra$^{1}$, and Sawyer B. Fuller$^{1}$
\thanks{This work was partly supported by the NSF award no. 2319060}
\thanks{$^{1}$ Aaron Weber, Daksh Dhingra, and Sawyer B. Fuller are with the Department of Mechanical Engineering, University of Washington, Seattle, USA
        {\tt\footnotesize aweber6@uw.edu, dd292@uw.edu, minster@uw.edu}}%
\thanks{A supplemental video for this paper is available here: https://www.youtube.com/watch?v=c35bT53Rl7Y}
\thanks{This work has been submitted to the IEEE for possible publication. Copyright may be transferred without notice, after which this version may no longer be accessible.}
}
\title{\LARGE \bf
A flexured-gimbal 3-axis force-torque sensor reveals minimal cross-axis coupling in an insect-sized flapping-wing robot
}
\begin{document}


\maketitle


\input{1_abstract.tex}
\input{2_introduction.tex}
\input{3_principle.tex}

\input{4_experiment.tex}

\input{5_results.tex}
\input{6_Conclusion.tex}


\bibliographystyle{IEEEtran}
\bibliography{citations}
\end{document}

%% file: 1_abstract.tex
\begin{abstract}

The mechanical complexity of flapping wings, their unsteady aerodynamic flow, and challenge of making measurements at the scale of a sub-gram flapping-wing flying insect robot (FIR) make its behavior hard to predict. Knowing the precise mapping from voltage input to torque output, however, can be used to improve their mechanical and flight controller design. To address this challenge, we created a sensitive force-torque sensor based on a flexured gimbal that only requires a standard motion capture system or accelerometer for readout. Our device precisely and accurately measures pitch and roll torques simultaneously, as well as thrust, on a tethered flapping-wing FIR in response to changing voltage input signals. With it, we were able to measure cross-axis coupling of both torque and thrust input commands on a 180 mg FIR, the UW Robofly. We validated these measurements using free-flight experiments. Our results showed that roll and pitch have maximum cross-axis coupling errors of 8.58\% and 17.24\%, respectively, relative to the range of torque that is possible. Similarly, varying the pitch and roll commands resulted in up to a 5.78\%  deviation from the commanded thrust, across the entire commanded torque range. Our system, the first to measure two torque axes simultaneously, shows that torque commands have a negligible cross-axis coupling on both torque and thrust. 

\end{abstract}

%% file: 2_introduction.tex
\section{Introduction}
Flying insect-sized robots (FIRs) are sub-gram robots that use flapping wings inspired by insects. Their small size and low weight gives them an advantage in terms of the ability to access places that are otherwise inaccessible by bigger drones.  For this reason, they have promising potential in applications like search and rescue missions, running inspections in manufacturing plants, and detecting gas leaks. 
Unlike birds and bats, bumblebees and other insects flap wings using a pair of thorax muscles. Inspired by biology, FIRs use a pair of piezoelectric actuators connected to the wings through a transmission system~\cite{perez2011first},~\cite{fuller2019four}, and ~\cite{beeplus}. In motors, friction forces and heat dissipation in coils increasingly dominate as the size gets smaller. Piezoelectric actuators can operate at high efficiency even at the centimeter scale. 

Single-input single-output control at the actuator level is important to achieve high level precision control in flying robots. Mahony et. al.~\cite{Mahonyquad2012} developed a motor model that converts the input PWM signal to the rotor speed for their quadrotor systems. Karasek et. al.~\cite{KarasekDelfly2018} used an electronic speed control with customized RPM sensing to achieve precise motor control on their 28g flapping-wing robot. However, modeling the output based on the input voltage in piezo-based systems is more challenging because: 1) piezoelectric actuators and the transmission systems used in the FIRs are manufactured and assembled by hand, resulting in greater manufacturing variability and more variable output thrust and output torque, and 2) FIR components are prone to high wear so even with the same input to the actuator, the output of a flapping wing changes over time. The flapping process puts high stress on the FIR and results in a short device lifespan \cite{malka2014principles}, and can cause the dynamics of the FIR to change during experimentation.

Control of these robots~\cite{chirarattananon2013adaptive}~\cite{fuller2019four}~\cite{ma2013controlled} has previously relied on the robustness of feedback control systems to compensate for uncertainties in command-to-output mapping. As a practical matter, however, approaching control in this way leads to significant amounts of trial and error in experiments to tune the gains of the controller specific to a robot. This not only reduces operator productivity, but it can also severely reduce the lifespan of the robot~\cite{malka2014principles}.

The goal of this work is to develop a method to simultaneously map the input signal of the piezoelectric actuators with the output thrust and torque of the robot. This would ultimately be used for writing more accurate controllers for FIRs and aid in characterizing the performance of new designs. 

Measuring torques at the scale output by FIRs is challenging because the small torques involved preclude using off-the-shelf sensor hardware. The smallest commercially available multi-axis torque sensor, the ATI Nano17 Titanium, has a resolution of 8~$\mu$Nm. This is an order of magnitude higher than what is needed to accurately measure FIR torques. One alternative that has previously been used to analyze the torque of FIRs is a capacitive torque sensor ~\cite{finio2011torques}. The downside of such systems is that they require expensive, specialized capacitive sensing hardware costing \$1000 or more. Furthermore, a two-axis version capable of measuring two torque axes at once has not yet been demonstrated at small scale. Here, we propose an alternative approach that, like capacitive and strain-gauge systems, allows the object to move by a small amount. In our system, however, the spring-like restoring torque is fairly low, so that the angular deflections are large enough to be detected by motion capture or inclinometer. By doing so, we are able to use hardware that is already available in many robotics contexts: either a camera-based motion capture system, or potentially an accelerometer. As a consequence, however, the bandwidth of our sensor is reduced, allowing it to only measure torques on a stroke-averaged basis. While we believe the torque or force changes nearly instantaneously in response to a voltage command, this has not been measured, and such dynamic measurements would require a different kind of sensor. 

To perform the desired measurements, we introduce a system that is conceptually similar to the device introduced in~\cite{ddhingraTrimming}, but that incorporates a number of improvements. These include flexure axes that now intersect the approximate center of mass of the vehicle, the addition of a damper to reduce unwanted oscillations,  the addition of a precision scale to measure forces, and precise calibration and validation to ensure that it can measure torque outputs produced by the FIR. 

This work is the first to perform simultaneous two axis torque mapping of FIRs. Until now, FIR controllers have assumed they have independent roll and pitch actuation, but this has never been measured directly. Our results show this assumption largely holds and demonstrate that torque actuation has a negligible impact on the upward thrust of the FIR.

%% file: 3_principle.tex
\section{Principle of Operation}
\begin{figure}[tbp]
    \centering
    \includegraphics[width=0.5\columnwidth]{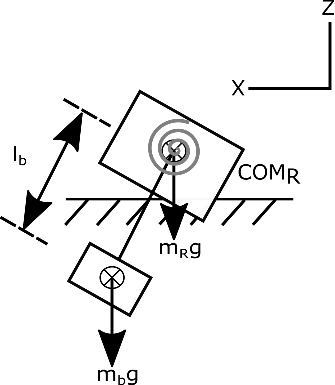}
    \caption{Principle of the torque measurement, showing the robot and the counterweight.}
    \label{fig:principle}
\end{figure}

\subsection{Torque measurement about flexure axis }

In our experiments we use a device introduced in~\cite{ddhingraTrimming} that constrains the robot to rotate around two axes- pitch and roll- while keeping all other degrees of freedom fixed. The axis is subject to a spring-like restoring torque so that the robot remains upright at equilibrium. When the wings are flapped, in general they produce both a thrust and a torque. The device is designed so that applied torque can be measured by the angular deflection while ensuring that thrust has no effect on rotation, regardless of magnitude.

Sensitivity of the device is the angular deflection per unit torque applied to the system. We want the minimum measurable torque by the device to be 0.3 $\mu$Nm. This is motivated by the estimated torque uncertainty induced by the thin wire tether that provides power and control signals to the robot. While it is hard to provide a simple model of its effect due to its widely variable conformation, one reasonable model for the tether is a torsional spring. Experiments performed in~\cite{fuller2014controlling} show that a 45$^{\circ}$ rotation causes a torque of approximately 0.3$\mu$Nm. Figure \ref{fig:principle} shows the principle behind such a device. 
A robot of weight $m_r$ and moment of inertia $I$ mounted on a torsional spring of flexure stiffness $K_f$ at a distance $l_r$ from the axis of rotation. A counterweight of mass $m_b$ is mounted at a distance of $l_b$ from the axis of rotation. New to this version of the device, a damping rod extends down into a dish of glycerin with damping coefficient $b$ to provide damping. Net torque about point O is:
\begin{align}
    \begin{aligned}
       \tau_{net} &= m_b l_b \sin\theta - m_R l_R \sin\theta + k_f \theta + b\dot\theta + I\Ddot{\theta}
    \end{aligned}
\end{align}  
As the new version of the device makes the robot axis of rotation level with the device axis of rotation, the $m_R l_R \sin\theta$ term can safely be ignored. For our application, the measurements will be taken when the robot is at a steady state, and as such the $\dot\theta$ and $\Ddot{\theta}$ terms will go to zero. Using the small-angle approximation,
\begin{align}
    \begin{aligned}
       k_s &= m_b l_b + k_f  
    \end{aligned}
    \label{equation_stiffness}
\end{align}
Where $k_s = \frac{\tau}{\theta}$ is the sensitivity of the device. Sensitivity can be increased by reducing the mass of the counterweight $m_b$. However, making the sensitivity too high could result in deflections that enter the non-linear zone of the elastic flexure joint.

\subsection{Actuating Torques}
The input to the FIR is a sinusoidal signal. The amplitude $V_{amp}$ of the sinusoidal signal is proportional to the thrust magnitude of each wing. Roll torques in flapping-wing platforms are produced by increasing the amplitude of one wing relative to the other wing, $ \Delta V=V_{amp1}-V_{amp2}$. A pitch torque is produced by altering the position of the
stroke-averaged center of aerodynamic thrust~\cite{ddhingraTrimming}. This is achieved by changing the mean of the sinusoidal signal by an offset voltage, $V_{off}$. We assumed the torques produced by the robot would be linear with respect to the inputs, which was validated by our results.


While mounted on the device when a robot generates torque it will produce angular acceleration, $\Ddot{\theta}$, angular velocity, $\dot{\theta}$, and angular deflection, $\theta$. However, in steady state conditions we can assume $\dot{\theta}$ and $\Ddot{\theta}$ terms to be zero. 

%% file: 4_experiment.tex
\section{Experimental Setup}

\subsection{Flexured-gimbal Device}
We introduce a new design with a few improvements over the device introduced in~\cite{ddhingraTrimming}. Like its earlier incarnation, our system has two independent axes of rotation. 

\begin{figure}[tbp]
    \includegraphics[width=\columnwidth]{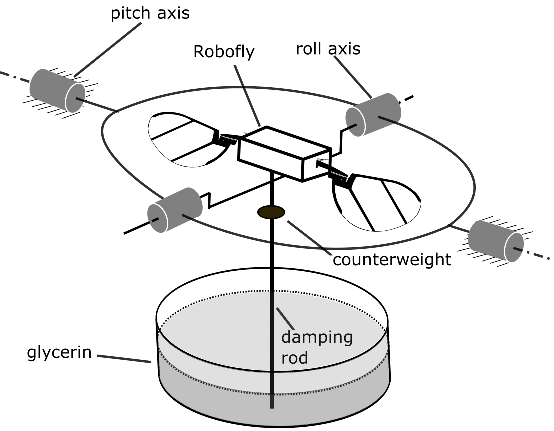}
    \caption{Diagram of the principle of operation of the flexure-based force/torque sensor. The flexures are positioned such that the roll and pitch axes of the device intersect, and intersect with the approximate center of mass of the FIR. The addition of a damping rod, below, whose end is immersed in glycerin, provides damping to eliminate oscillations.}
    \label{axisFig}
\end{figure}

The first improvement is that our new system moves the axes of rotation of the roll and pitch degree of freedom so that they both now intersect the approximate location of the center of mass of the FIR (Figure \ref{axisFig}.) In the earlier design, one of the flexure axes was well below the FIR. This improvement means that now, thrust force force from the wings that is not perfectly vertical, which could happen if the FIR is slightly tilted, no longer causes a torque disturbance in measurements. We made this design change by enlarging the circular gimbal shape so that it encircled the flapping wings with enough distance to avoid impeding flow.  

Second, the new system adds a dish of glycerin below to attenuate undesirable resonant oscillations of the system. 

And third, it is mounted on a precision balance so that forces can be measured simultaneously with torques. This can be seen in the supplementary video~\cite{robotVideo}.

The joints of the system are made of a flexible layer of 12~$\mu$m Kapton sandwiched between two 254~$\mu$m fiberglass (FR4) layers. Machining the larger gimbal shape required performing two separate cuts with the galvo-steered laser, with a precise stage move in between. This was because was too large to fit within the $\approx$ 50~mm  cutting area of the galvo. The two fiberglass layers were then bonded with the Kapton layer using Pyralux adhesive sheets. 

\begin{figure}[tbp]
    \includegraphics[width=\columnwidth]{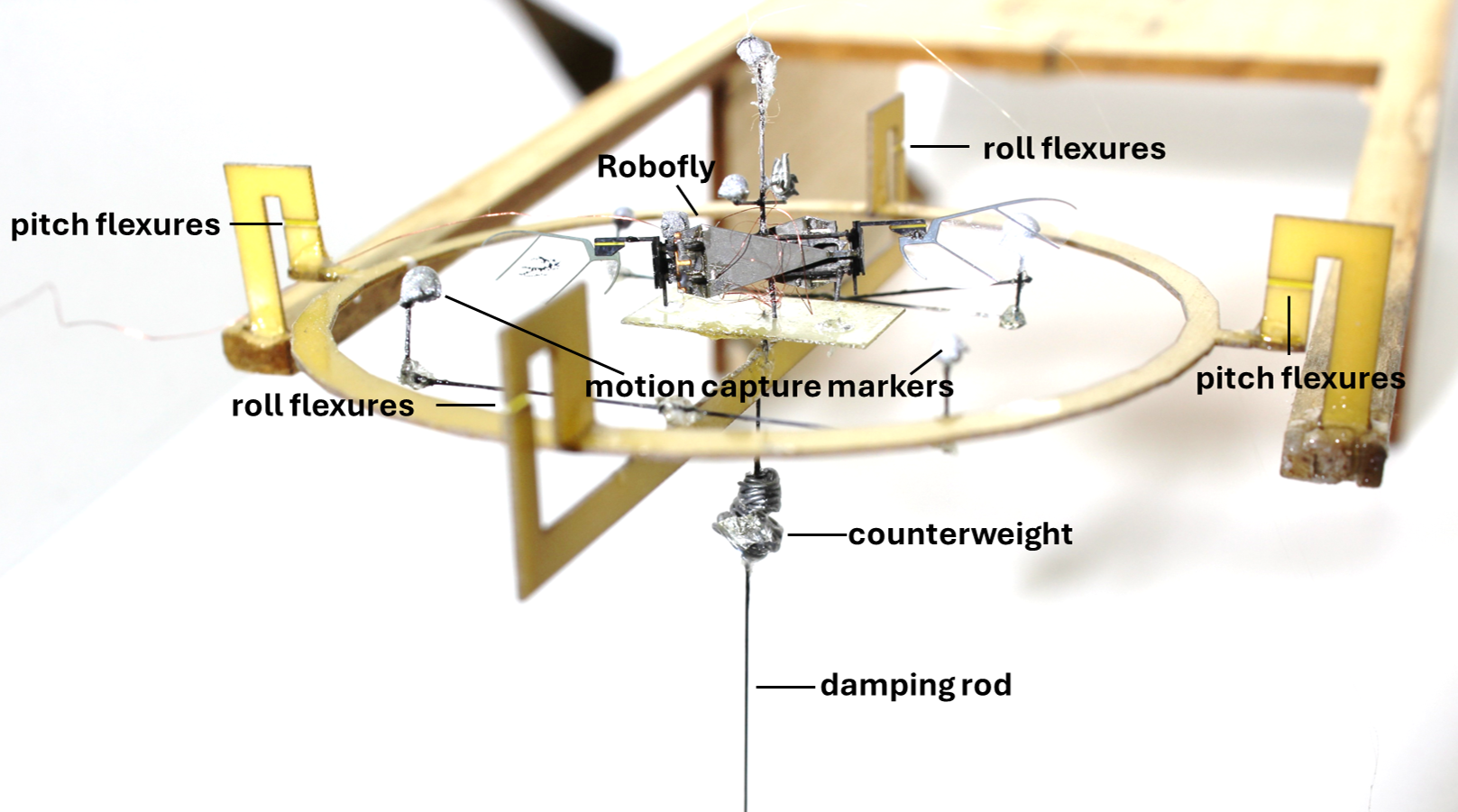}
    \caption{Image of the flexured-gimbal device with a FIR attached. Not shown in the figure is the glycerin petri dish below. Readout is accomplished in this case using a camera-based motion capture system.}
    \label{deviceLabels}
\end{figure}

To fabricate the device, we aligned all the layers using tight-fit pins and pressed it under 50~kgf force and 200$^{\circ}$C  temperature to adhere them. The base joints were then mounted on a platform that positions the robot at a height of 65 mm to minimize the ground effect. Once the device is assembled, a stage is attached and set slightly below the flexure joints, so that when the FIR is mounted its roll and pitch axes of rotation will be level with the flexures of the device. Additionally, a thin rod is attached extending below the stage, which sits in a petri dish filled with glycerin for damping. The device can be seen in Figure \ref{deviceLabels}.

\subsection{UW Robofly}

For the experiments we are using the FIR introduced in~\cite{ChukewadTRO}. The robot weighs 180~mg (including motion capture markers) and flaps its wings at 180 beats per second. 

\subsection{Performing  measurements}
A crucial step before data collection is aligning the center of rotation of the robot with the axis of rotation of the flexures on the gimbal. As done in~\cite{ddhingraTrimming}, this calibration procedure involved moving the robot laterally/longitudinally on the device until its pitch and roll angles are equal to what they were before the robot was added. 
Once the FIR has been mounted on the device, the whole gimballed system was placed on the scale so that thrust values may also be measured. The base of the device has been redesigned from the previous iteration to allow the base to sit on the scale while the FIR hangs out the side off of the measurement platform, to avoid measurement errors. During the testing period, a variety of control signals are sent to the FIR, with different values of $V_{off}$ and $\Delta V$, and the angle values are measured with the motion capture system and the angle data is taken from the steady state portion of each flight. This data can then be used to determine the torque-voltage mapping. 


Two FIRs, the ``mapping fly" and the ``validation fly," were used for experimental data collection. The mapping fly received various control signals for mapping analysis, but the stress from this process made it unusable for further testing. Consequently, the validation fly was used for a shorter mapping process. "Ground truth" measurements were obtained on the validation fly by applying static free-flight torques with small offset weights on rods extending from the robot's center and then trimming it in free flight. Using two separate robots with different actuator dynamics, trims, and manufacturing uncertainties demonstrates the generality of our results and shows uniformity and accuracy of our proposed method.

%% file: 5_results.tex
\section{Results}

\subsection{Flexured-gimbal Device Calibration}

\begin{figure}[tbp]
    \includegraphics[width=.9\columnwidth]{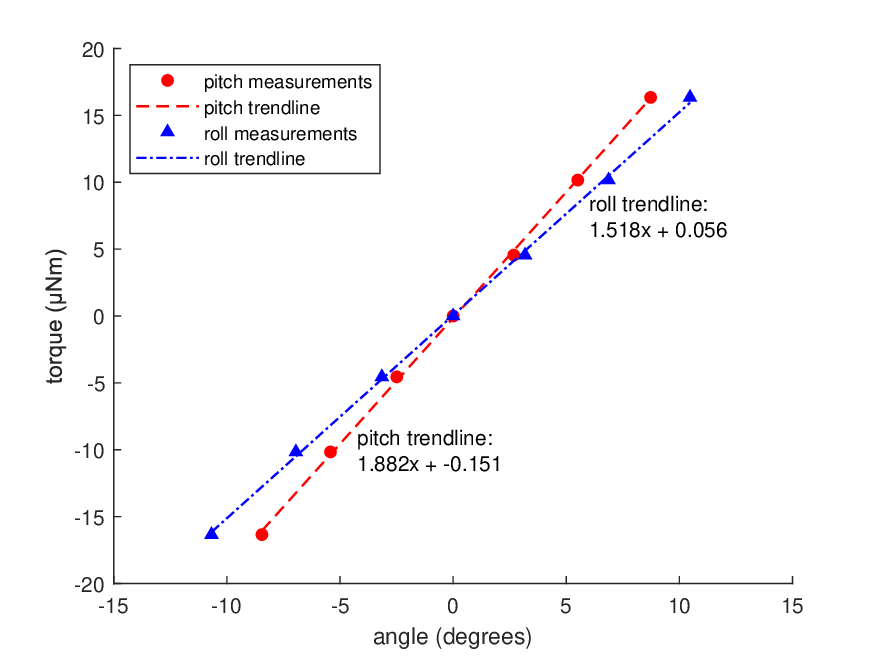}
    \caption{Flexured-gimbal device sensitivity measurements in the roll and pitch axes, with calculated trend lines.}
    \label{calibration_fig}    
\end{figure}

To calculate the sensitivity of the flexure joints, we applied known torques about the axis of rotation and noted the resulting static angular deflections in the flexure joints. The torques are applied in both clockwise and counterclockwise directions by hanging known masses at some known distances from the axis of rotation. 
Figure \ref{calibration_fig} shows the data points of static angular deflections from applied torques about the roll and pitch axis. The slope of these lines is the sensitivity of the flexure joints. For our device, the sensitivity was found to be 1.518~$\mu$Nm/rad in roll and 1.882~$\mu$Nm/rad in pitch, using a least-squares fit.

\subsection{Trimming Results}
To establish a starting point for the range of torque mapping measurements, the FIRs were trimmed in free flight before being attached to the device to find the values of $\Delta V$ and $V_{off}$ (as described in section II) that allow for a straight takeoff while canceling out undesired bias roll and pitch torques from manufacturing errors. The initial mapping fly successfully took off with the trimming values of 34~V roll trim ($\Delta V$) and $-3$~V pitch trim ($V_{off}$). The validation fly successfully took off with the trimming values of $-20$~V roll trim and 7.5~V pitch trim. For the validation fly offset torques, a 31.8~mg mass was added on a rod 4~mm from the center along the roll axis, resulting in a 1.248~$\mu$Nm roll torque, and the adjusted roll trim required for takeoff was $-17$~V. Similarly, a 25~mg mass was added on a rod 4mm from the center along the pitch axis, resulting in a $-0.981$~$\mu$Nm pitch torque, and the adjusted pitch trim required for takeoff was 5~V. After the mapping measurements the validation fly was trimmed one final time, and took of with values of $-23$~V roll trim and 5~V pitch trim.

\subsection{Torque Mapping}
After trimming in free flight, the FIR was mounted on the device and angle measurements were taken at varying roll and pitch voltages. The pitch voltages used were 0~V, ±5~V, ±10~V, and ±15~V, and the roll voltages used were 0~V, ±15~V, ±30~V, and ±50~V. All combinations of the listed roll and pitch voltages were tested, with the exception of ±15~V pitch and ±50~V roll, as these high roll and pitch values were hitting the actuator limits. These limits are set by the bias voltage of 250~V and the analog sinusoidal signal floats between 0 and 250. When the pitch and roll goes higher than ±15 and ±50 respectively, the flapping signal hits the limits of 0 and 250~V. Using the flexured-gimbal device and motion capture setup, the control signal was sent to the FIR on the device and its roll and pitch angles were measured. The control signal was sent for 3 seconds, and the measurements from the last 0.5 seconds were taken and averaged for use in the torque mapping, to allow the FIR to reach a steady angle on the device.

\begin{figure}[tbp]
    \begin{subfigure}{\columnwidth}
        \includegraphics[width=\columnwidth]{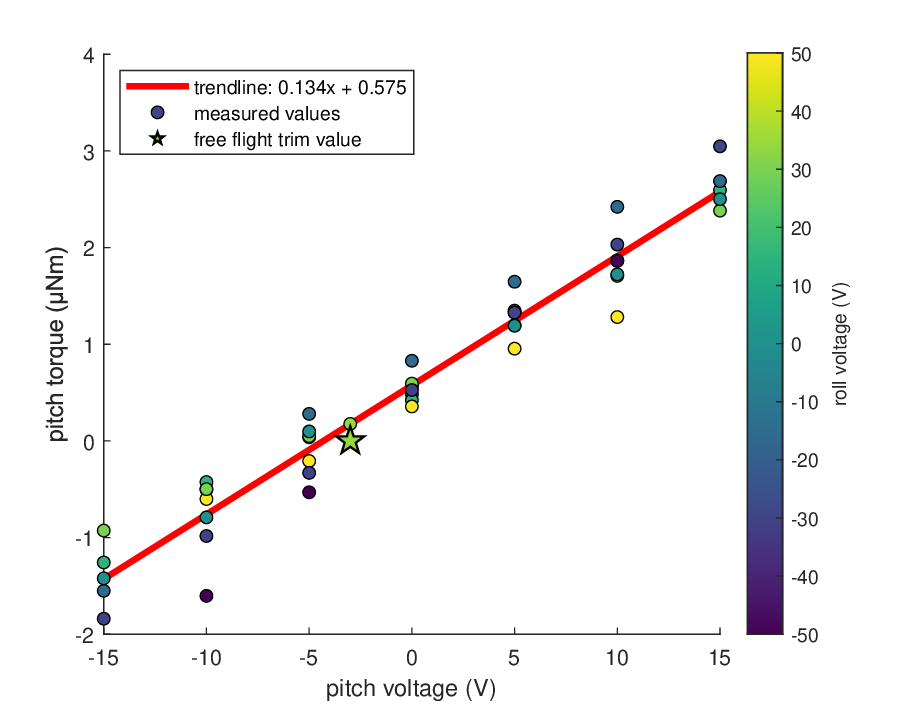} 
        \caption{ }
        \label{pitch1}
    \end{subfigure}
    \begin{subfigure}{\columnwidth}
        \includegraphics[width=\columnwidth]{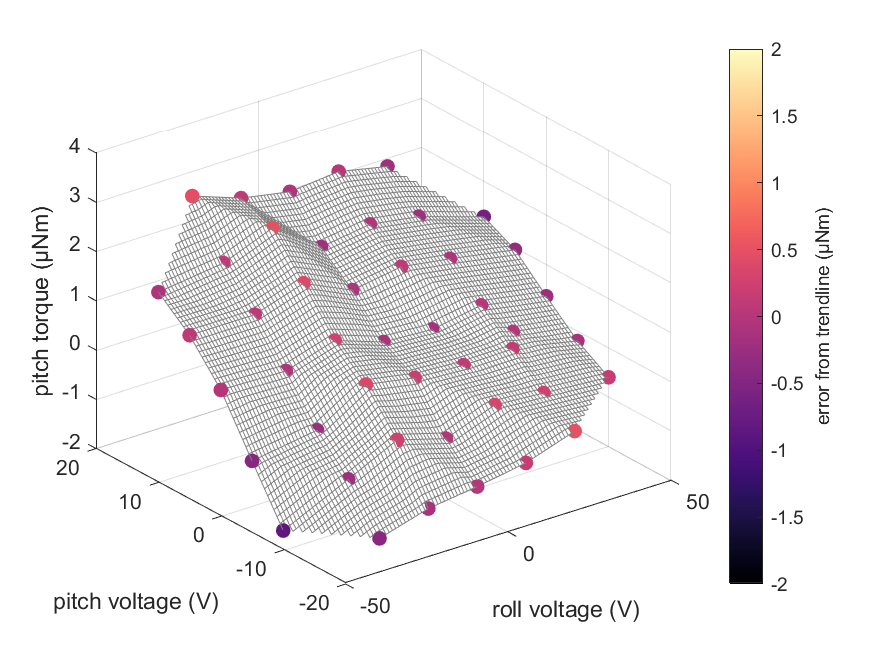}
        \caption{ }
        \label{pitch3d}
    \end{subfigure}
    \caption{(a). Mapping of the pitch voltage offsets in the control signal to the resulting pitch torque measured by the device, with a color map to show the strength of the roll voltages at each data point. (b). Mapping of the roll and pitch voltage offsets in the control signal to the resulting pitch torque measured by the device. Error from the mapping trendline is shown via the colormap at the measurement points. Pitch torque is not significantly impacted by changes in roll control voltage.}
    \label{pitchCombo}
\end{figure}

\subsubsection{Pitch measurements}
Shown in Figure \ref{pitch1} is the mapping of the pitch control voltages to the resulting pitch torques calculated using the measured angles and the angle-to-torque mapping found in the flexured-gimbal device calibration section. Multiple measured values are shown at the different values of pitch voltages due to the same pitch voltage being used with multiple different roll voltages. The mapping can be fit to a linear trendline, which fits the measured data with a 0.95 coefficient of determination. 

\begin{figure}[tbp]
    \begin{subfigure}{\columnwidth}
        \includegraphics[width=\columnwidth]{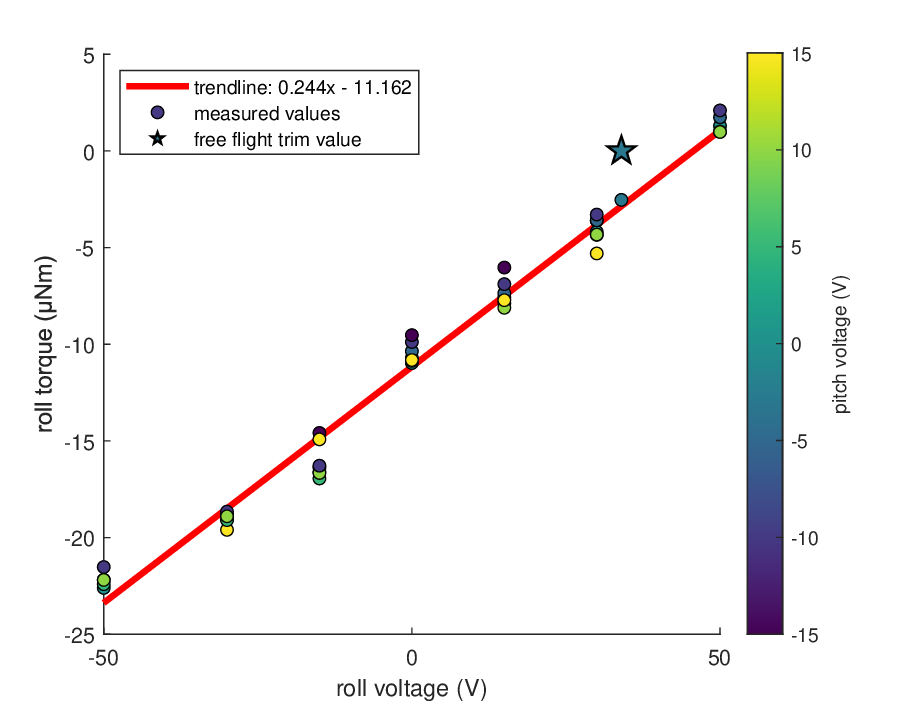} 
        \caption{ }
        \label{roll1}
    \end{subfigure}
    \begin{subfigure}{\columnwidth}
        \includegraphics[width=\columnwidth]{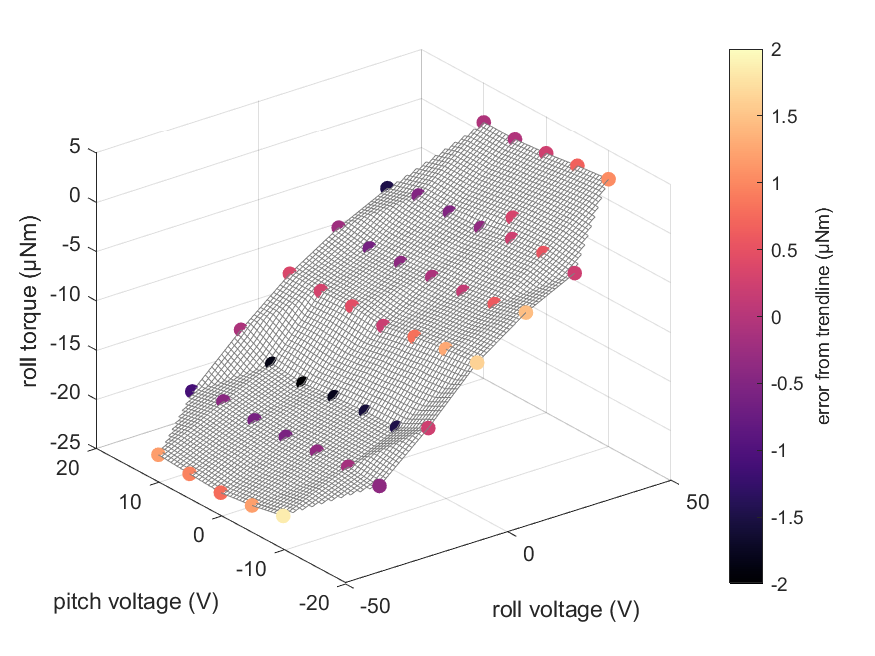}
        \caption{ }
        \label{roll3d}
    \end{subfigure}
    \caption{(a). Mapping of the roll voltage offsets in the control signal to the resulting roll torque measured by the device, with a color map to show the strength of the pitch voltages at each data point. (b). Mapping of the roll and pitch voltage offsets in the control signal to the resulting roll torque measured by the device. Error from the mapping trendline is shown via the colormap at the measurement points. Roll torque is not significantly impacted by changes in pitch control voltage.}
    \label{rollCombo}
\end{figure}
\begin{figure}[tbp]
    \begin{subfigure}{\columnwidth}
        \includegraphics[width=\columnwidth]{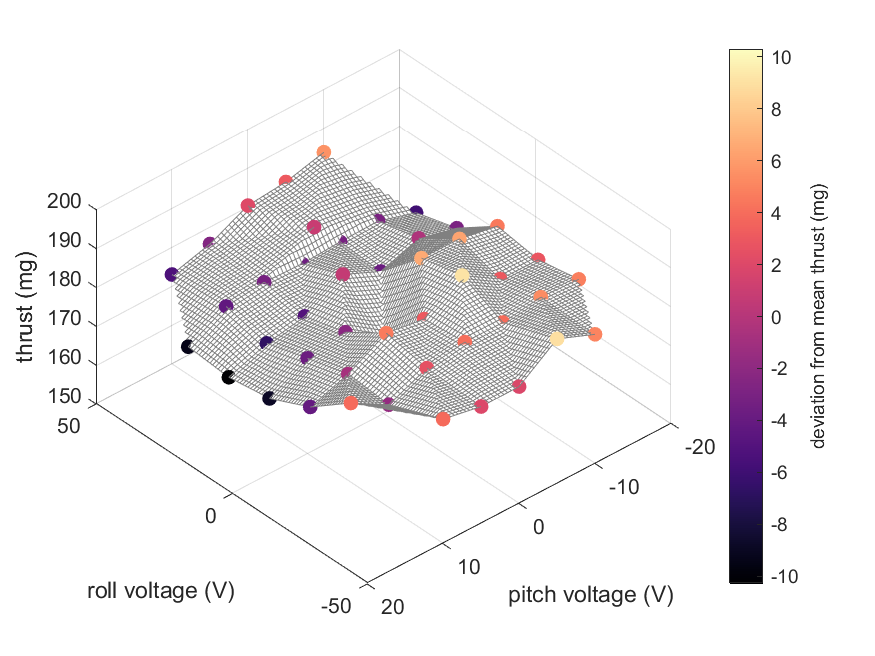} 
        \caption{ }
    \end{subfigure}
    \begin{subfigure}{\columnwidth}
        \includegraphics[width=\columnwidth]{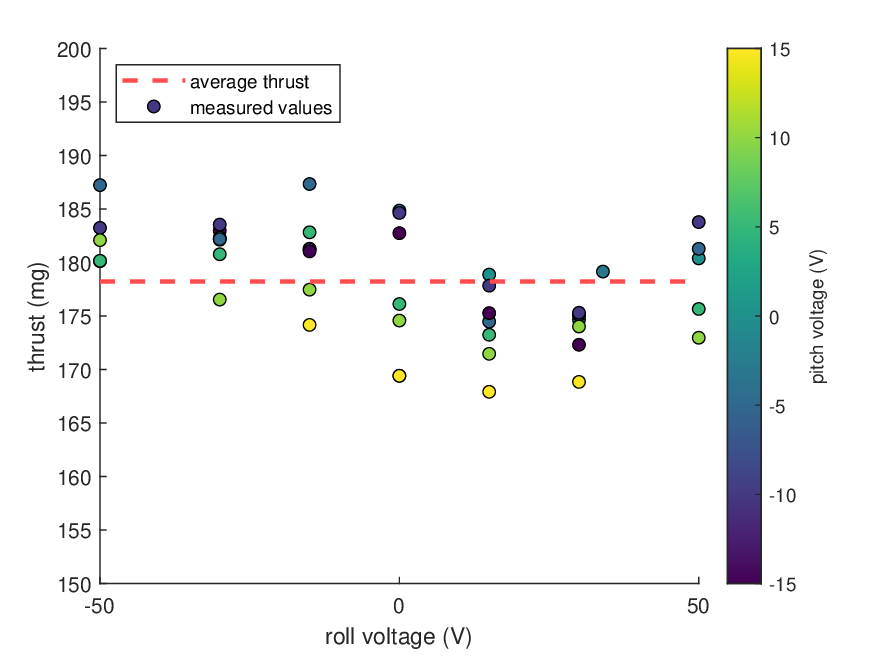}
        \caption{ }
    \end{subfigure}
    \caption{(a). Mapping of the roll and pitch voltage offsets in the control signal to the resulting FIR thrust, with a colormap indicating deviation from the mean thrust value. (b). Mapping of roll voltage to resulting FIR thrust with a colormap indicating the corresponding pitch voltages, showing that the thrust is not significantly impacted by the roll voltage but that there is a slight trend of lower thrust values corresponding to higher pitch voltages.}
    \label{thrustFig}
\end{figure}

\subsubsection{Roll measurements}
Shown in Figure \ref{roll1} is the mapping of the roll control voltages to the resulting roll torques, calculated in the same manner as the pitch mapping. As with pitch, multiple measured values at the same roll voltage are where the same roll voltage was used with different pitch voltages. As with the pitch mapping, the roll mapping can be fit to a linear trendline, which fits the measured data with a 0.98 coefficient of determination.

\subsubsection{Pitch and Roll Torque Coupling}
In addition to mapping the roll and pitch control voltages to the resulting roll and pitch torques, we also wanted to measure if roll voltage had a significant effect on the pitch torque, and if pitch voltage had a significant effect on the roll torque. The results are shown in Figures \ref{pitch3d} and \ref{roll3d}, with error from the trendline indicated by the color mapping. Larger error values are mostly on the edges of the plots where the control voltages are higher,likely due to these control voltages being at the upper limit of the signal range and not representative of typical flight conditions. The cross-axis coupling error was calculated as the offset between the measured torque and the expected torque from the trendline, relative to the total torque range actuated by the FIR. At the points with the most extreme control voltage in the opposite axis, the maximum error in the torque mapping is 2.12~$\mu$Nm in the roll axis, which gives a percentage error of 8.58\% relative to the total roll torque actuated, and .84~$\mu$Nm in the pitch axis, which gives a percentage error 17.24\% error relative to the total pitch torque actuated.

\subsubsection{Relation between thrust and control voltages}
The flexured-gimbal device was mounted on a scale to measure the thrust of the FIR during the trials. As seen in Figure \ref{thrustFig}, there is some variation in the thrust with the varying control voltages, but the maximum deviation from the mean thrust is small (5.78\%) and there is only a weak trend connecting roll and pitch voltages to thrust (a slope of $-0.257$ mg/V in pitch and $-0.078$ mg/V in roll).

\subsection{Validation}
\begin{figure}[tbp]
    \includegraphics[width=\columnwidth]{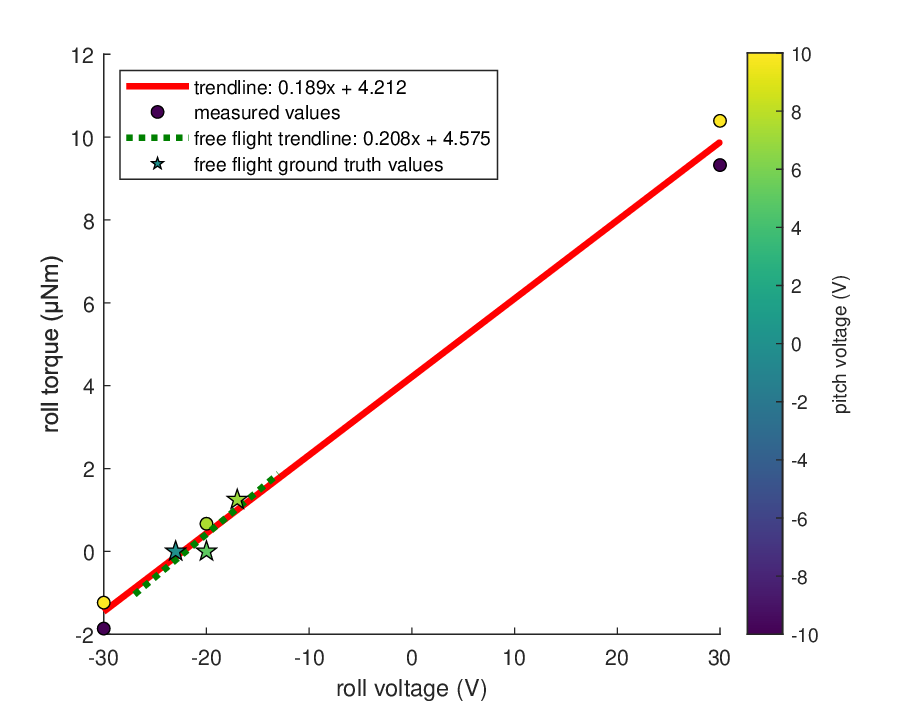}
    \caption{Control voltage to output torque mapping results from the device for the roll torque of the validation fly, along with the roll torque values mapped using free flight experiments with offset torques.}
    \label{rollVal}
\end{figure}

\begin{figure}[tbp]
    \includegraphics[width=\columnwidth]{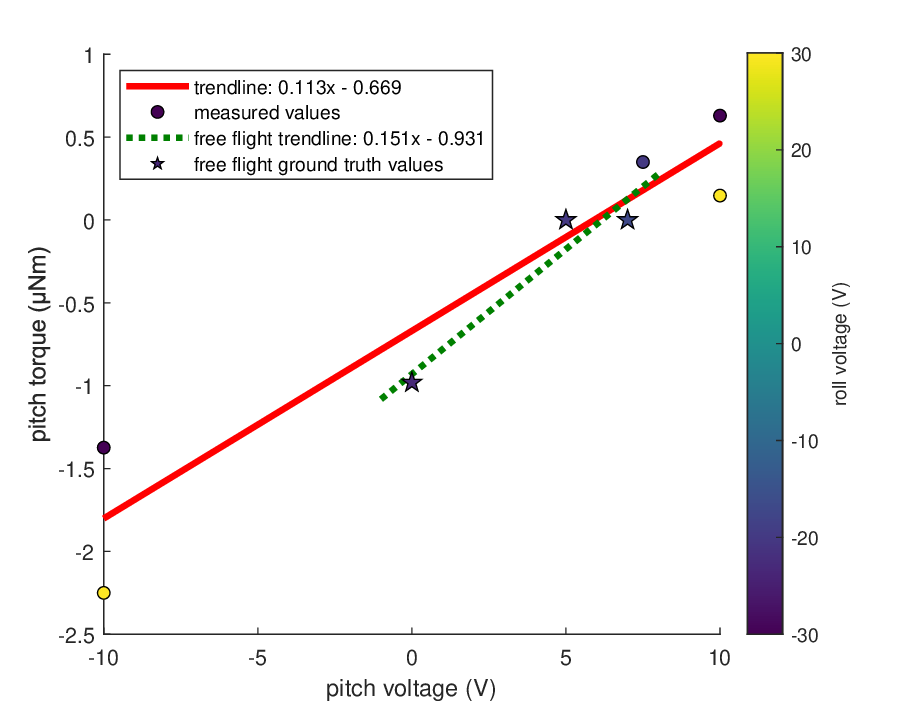}
    \caption{Control voltage to output torque mapping results from the device for the pitch torque of the validation fly, along with the pitch torque values mapped using free flight experiments with offset torques.}
    \label{pitchVal}
\end{figure}

Only five measurements were taken in the mapping stage with the validation fly, to produce a large enough range to develop a mapping trendline fit while preserving the lifespan of the FIR and reducing wear. Following the mapping measurements the FIR was removed from the device, then trimmed in free flight again. Notably, the trim values changed slightly due to wear even with the shorter mapping process, resulting in two different control voltage data points for zero torque in free flight (shown in Figures \ref{rollVal} and \ref{pitchVal}). Finally, the FIR was fitted with weights providing torque offsets in pitch and roll and trimmed in free flight, so that the control voltages needed to provide the torque counteracting the offsets could be calculated. Figures \ref{rollVal} and \ref{pitchVal} show the measured data, as well as trend lines for the free flight experiment mapping and the device mapping. The measurement error of the device relative to the ``ground truth" free-flight experiments was 9\% for roll and 25\% for pitch. The larger percentage error in pitch is due to the smaller size of torque being measured relative to the disturbance. The largest error in Figure~\ref{pitchVal}, occurring at 0~V pitch, is about equal to the torque disturbance of the wire tether measured in~\cite{fuller2014controlling}  of 0.3~$\mu$Nm. 

%% file: 6_Conclusion.tex
\section{Conclusion and Future Work}

This paper reports a device design and measurement process that can be used to map control voltages to the resulting roll and pitch torques produced in a very small flapping-wing robot, even below a gram. This system is an improvement over an earlier system~\cite{ddhingraTrimming}  that was only capable of finding compensatory trim values rather than measuring torques directly. Like that system, the system here is constructed entirely using parts that are likely available in a lab or factory creating FIRs. Its gimbal and flexures are machined using the same laser system used to construct the robot itself. Readout is performed using a motion capture system that is standard equipment in many robotics settings. As a consequence of these choices, however, the device is limited to low-frequency measurements with bandwidth of approximately 0.3 Hz.

We found that the roll and pitch of the flying insect robot (FIR) we tested, the 180 mg UW Robofly, are decoupled and therefore can be actuated independently. This finding is consistent with the assumption that has been used to date in the design of the flight controllers of two-winged FIRs, which is that cross-axis coupling of torque commands is negligible. We anticipate this new information can be used to better model the dynamics of flapping-wing robots and control their movements more effectively, especially when undergoing aggressive (high-torque) maneuvers. The device is simple to construct and can be easily adapted for use with other types of flapping-wing robots. 

Further improvements can potentially be made to the device to increase the ease of use by using an accelerometer as an inclinometer to measure the angles instead of requiring a motion capture system. This  would simplify its use. It is expected most if not all fully-autonomous FIRs will have an accelerometer as an integral part of their inertial navigation system. Preliminary work, however, indicates that vibrations due to the flapping wings produce too much noise to recover the angle. These vibrations could potentially be attenuated by placing the accelerometer on the damping rod, down near the glycerin. 